\documentclass{article}
\usepackage{ijcai24}

\usepackage{times}
\usepackage{soul}
\usepackage{url}
\usepackage[hidelinks]{hyperref}
\usepackage[utf8]{inputenc}
\usepackage[small]{caption}
\usepackage{subcaption}
\usepackage{graphicx}
\usepackage{amsmath}
\usepackage{amsthm}
\usepackage{booktabs}
\usepackage{algorithm}
\usepackage{algorithmic}
\usepackage{tabularx}
\usepackage[switch]{lineno}

\usepackage{xcolor}

\title{Real-time Multi-modal Object Detection and Tracking on Edge for Regulatory Compliance Monitoring}
\author{
Jia Syuen Lim$^{1,2}$\footnote{This work was done when Jia Syuen Lim was an intern at CSIRO, Australia}
\and
Ziwei Wang$^2$\and
Jiajun Liu$^{2,1}$\and
Abdelwahed Khamis$^2$\and\\
Reza Arablouei$^2$\and
Robert Barlow$^2$\And
Ryan McAllister$^2$\\
\affiliations
$^1$The University of Queensland, Australia\\
$^2$CSIRO, Australia\\
\emails
jiasyuen.lim@uq.edu.au,\\
\{ziwei.wang, jiajun.liu, abdelwahed.khamis, reza.arablouei, robert.barlow, ryan.mcallister\}@csiro.au
}
\begin{document}
\maketitle

\begin{abstract}
    Regulatory compliance auditing in agrifood processing facilities is crucial for upholding the highest standards of quality assurance and traceability. However, the current manual and intermittent auditing approaches present significant challenges and risks, potentially leading to gaps or loopholes in the system. To address these shortcomings, we introduce a real-time, multi-modal sensing system. This system utilizes 3D time-of-flight and RGB cameras while leveraging unsupervised learning techniques on edge AI devices. The proposed system enables continuous object tracking, improving record-keeping efficiency and reducing the cost of manual labor. We demonstrate the effectiveness of the system in a knife sanitization monitoring scenario, showcasing its capability to overcome occlusion and low-light performance limitations commonly encountered with conventional RGB cameras.
\end{abstract}

\section{Introduction}

Regulatory compliance auditing of food processing facilities is a mandatory requirement in the agrifood export industries. However, the majority of these audits are currently conducted through manual inspections, which are intermittent and lack traceability. For instance, crucial hygiene regulations such as handwashing \cite{DBLP:conf/sensys/KhamisKCMH20} \cite{DBLP:conf/ipsn/KhamisKCMH20} and knife sanitizing, which are essential for preventing food-borne illnesses and cross-contamination, are often monitored manually, leading to potential inconsistencies or oversight. Implementing automated regulatory compliance checkpoints, along with sensor-driven data collection and machine learning-based analytics, can offer a robust solution. It can not only reduce labor costs significantly but also enhance quality assurance by enabling continuous monitoring and systematic record-keeping.

With the growing computational power of modern edge AI devices, the deployment of deep neural networks (DNNs) for real-time detection of visual objects has become more prevalent~\cite{DBLP:conf/iccvw/StackerFHBRSS21}. However, most existing DNN-based algorithms in computer vision applications are limited to two-dimensional images and lack the ability to incorporate geometrical knowledge or fine details.
Their performance can degrade due to occlusion and low ambient light settings.
Recent years have witnessed remarkable advancements in compact yet high performance 3D time-of-flight (ToF) technology.
The increased accessibility of ToF technology has facilitated the creation of accurate geometrical and spatial representations of physical objects in complex environment, including low-light conditions.
By integrating information from multiple sensing modes and leveraging their complementary properties, the drawbacks of perception and blind spots associated with individual modes can be effectively mitigated. 
This integration can lead to improved accuracy and robustness in real-world applications, such as behavior recognition~\cite{DBLP:conf/mobicom/WangLABMB22}.

\begin{figure*}[t]
  \centering
  \includegraphics[height=15em]{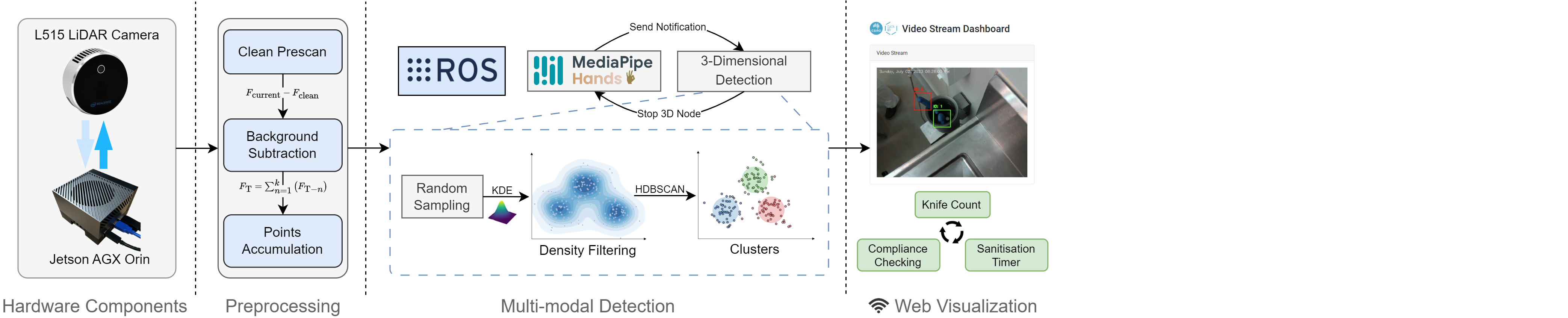}
  \caption{The overall architecture of our proposed system.}
  \label{fig:overview}
\end{figure*}

In this work, we introduce a novel near-real-time multi-modal sensing system designed specifically for object tracking on edge devices. 
It employs an integrated ToF camera comprising a 3D ToF sensor and a RGB camera. The inclusion of additional depth information enables precise identification of the static environment and moving objects.
We evaluate the proposed system considering a knife sanitization monitoring application, aiming to develop a knife detection and tracking model that eliminates the need for manual annotation.
The developed prototype employs an unsupervised multi-modal detector to estimate knife locations within a knife sterilizer, irrespective of different positioning and color settings. The performance of the system is evaluated in a laboratory setting that simulates a real meat processing factory environment.
We describe the proposed system in section~\ref{sec:system} and present some performance evaluation results in section~\ref{sec:results}.

\section{System Design}\label{sec:system} 
The proposed near-real-time detection and tracking system, depicted in Figure~\ref{fig:overview}, comprises two main hardware components: an integrated ToF camera and an edge AI device. We utilize the Robot Operating System 2 to manage multi-modal data streams from the ToF camera to the edge AI device. This system setup allows us to preprocess data and implement our detection and tracking pipeline in near real-time.

\subsection{Hardware Components}

\subsubsection{Time-of-Flight Camera} We utilize an Intel® RealSense™ ToF camera, which is a compact and lightweight device equipped with high-resolution ToF sensing technology, which is ideal for edge deployment. We initialize both its RGB camera and ToF depth sensor to operate at a frame rate of 30Hz.

\subsubsection{Edge AI Device} We utilize an NVIDIA® Jetson AGX Orin™ development kit to receive and process data captured by the ToF camera via USB-C connection.

\subsection{Software Pipeline}
\subsubsection{Robot Operating System 2 (ROS 2)} 
We make use of the open-source middleware ROS 2, which provides a range of services including device control, message-passing, and package management for robotics software.

\subsubsection{Preprocessing Components} 
A series of preprocessing steps are implemented prior to detection modules.

\paragraph{a) Static Environment Prescan.}
Initially, we conduct a comprehensive prescan of the static environment, ensuring that the knife sterilizer is clear of any objects, and save this scan as the static background model.

\paragraph{b) Background Subtraction.} We apply a coarse-to-fine filtering process to subtract the static background from the point cloud. First, to minimize the computational load, we apply a coarse filter to exclude points that fit within the voxel grid of background model. Next, we enhance the background subtraction process by employing a fine filter that utilizes a KDTree search to recursively isolate points within a specific distance from the background, assuming that distant points belong to foreground objects.

\paragraph{c) Accumulation of Points.} We implement a first-in-first-out (FIFO) queue mechanism to accumulate point clouds from $k$ successive frames. This approach allows for the generation of more comprehensive and denser point clouds.

\subsubsection{3D Object Detection}
Our 3D detection pipeline can be divided into several components as in the following.

\paragraph{a) Random Sampling.} The uneven distribution of accumulated points can introduce noise, potentially compromising the performance of our algorithm. To address this issue, we employ a random subsampling technique to effectively reduce the unevenness and noise in the point distribution, whilst enhancing the overall efficiency.

\paragraph{b) Gaussian Kernel Density Estimation (KDE).} Traditional clustering techniques often encounter difficulties when dealing with closely clumped point clouds, which is frequently observed in scenarios where knives are densely packed. To address this, we employ the KDE algorithm to predict the density map of the point clouds and employ percentile filtering to enhance the separability of the point clouds.

\paragraph{c) Hierarchical DBSCAN.} After processing the point cloud by the KDE algorithm, we use the hierarchical density-based spatial clustering of applications with noise (HDBSCAN) \cite{DBLP:journals/jossw/McInnesHA17} algorithm to determine the number of knives present.

\paragraph{d) Simple online and realtime tracking (SORT).} SORT \cite{DBLP:conf/icip/BewleyGORU16} is a real-time algorithm designed to track multiple objects in a live video stream. We obtain the 3D cluster centroids from HDBSCAN and project them onto a 2D image plane. The projected centroids are then fed to SORT for subsequent 2D object tracking.

\subsubsection{Hand Detection}
During the knife counting process, the knife handles may occasionally be occluded by workers' hands when they enter the scene. We use the MediaPipe \cite{DBLP:journals/corr/abs-2006-10214} Hand Landmarker tool to detect the presence of hands in live image streams. When a hand is detected, an event notification is dispatched to the 3D detection node, instructing it to temporarily halt the knife counting process until no hand is detected.

\section{Experiment and Results}\label{sec:results}
To assess the performance of the proposed system, we conducted an experiment at a pilot plant within CSIRO, under the approval of the relevant ethics committee. During the experiment, a ToF camera was set up on a tripod to capture a top-down view of a knife sterilizer. We designed our experiment to simulate a range of real-world scenarios. In total, we established five different scenarios, each representing a different number of knives, from one to five. For each scenario, we created 50 unique instances, representing various knife positions within the sterilization bath.

\begin{figure}[!b]
     \centering
     \begin{subfigure}[b]{0.155\textwidth}
         \centering
         \includegraphics[width=\textwidth]{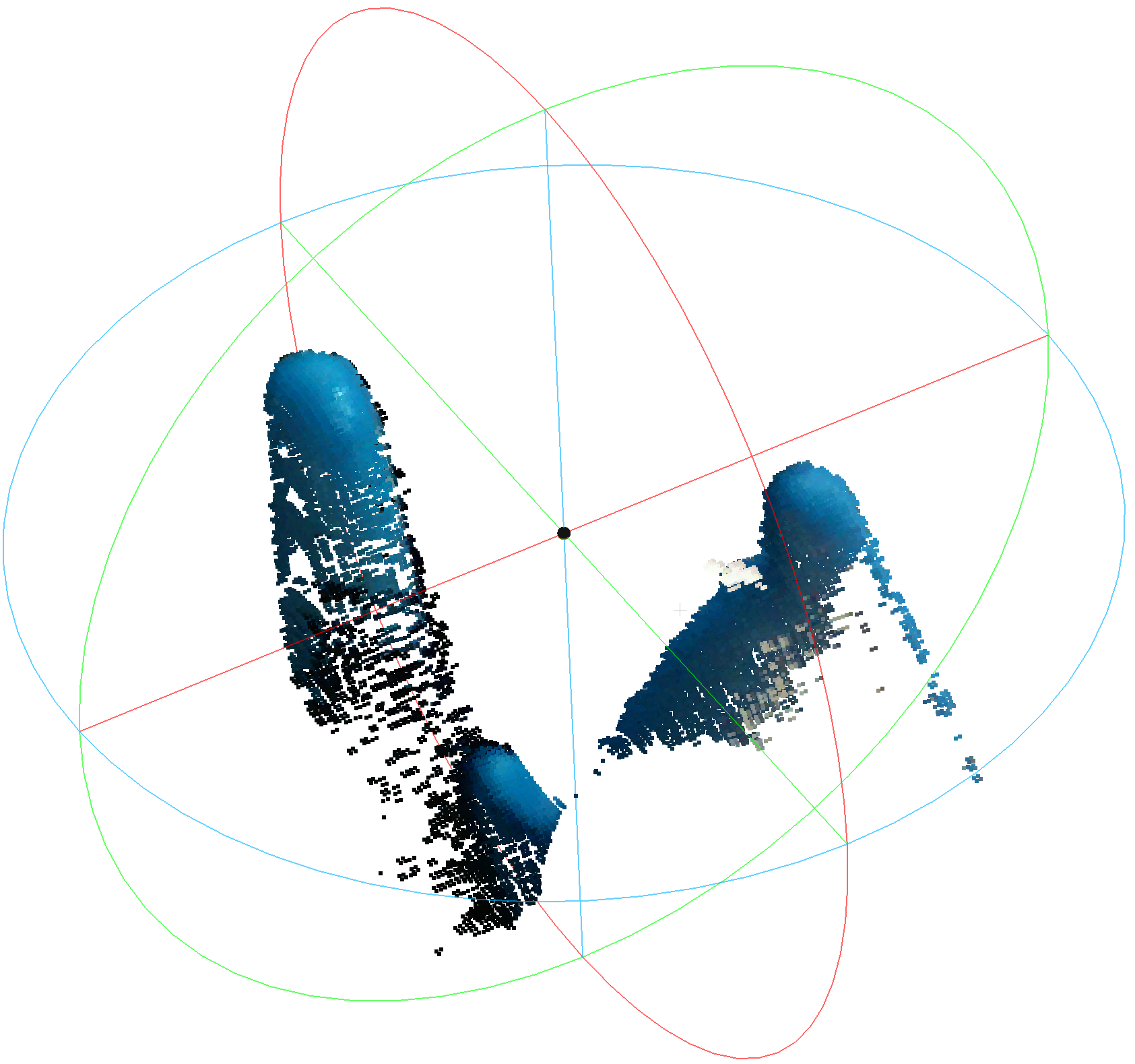}
     \end{subfigure}
     \begin{subfigure}[b]{0.155\textwidth}
         \centering
         \includegraphics[width=\textwidth]{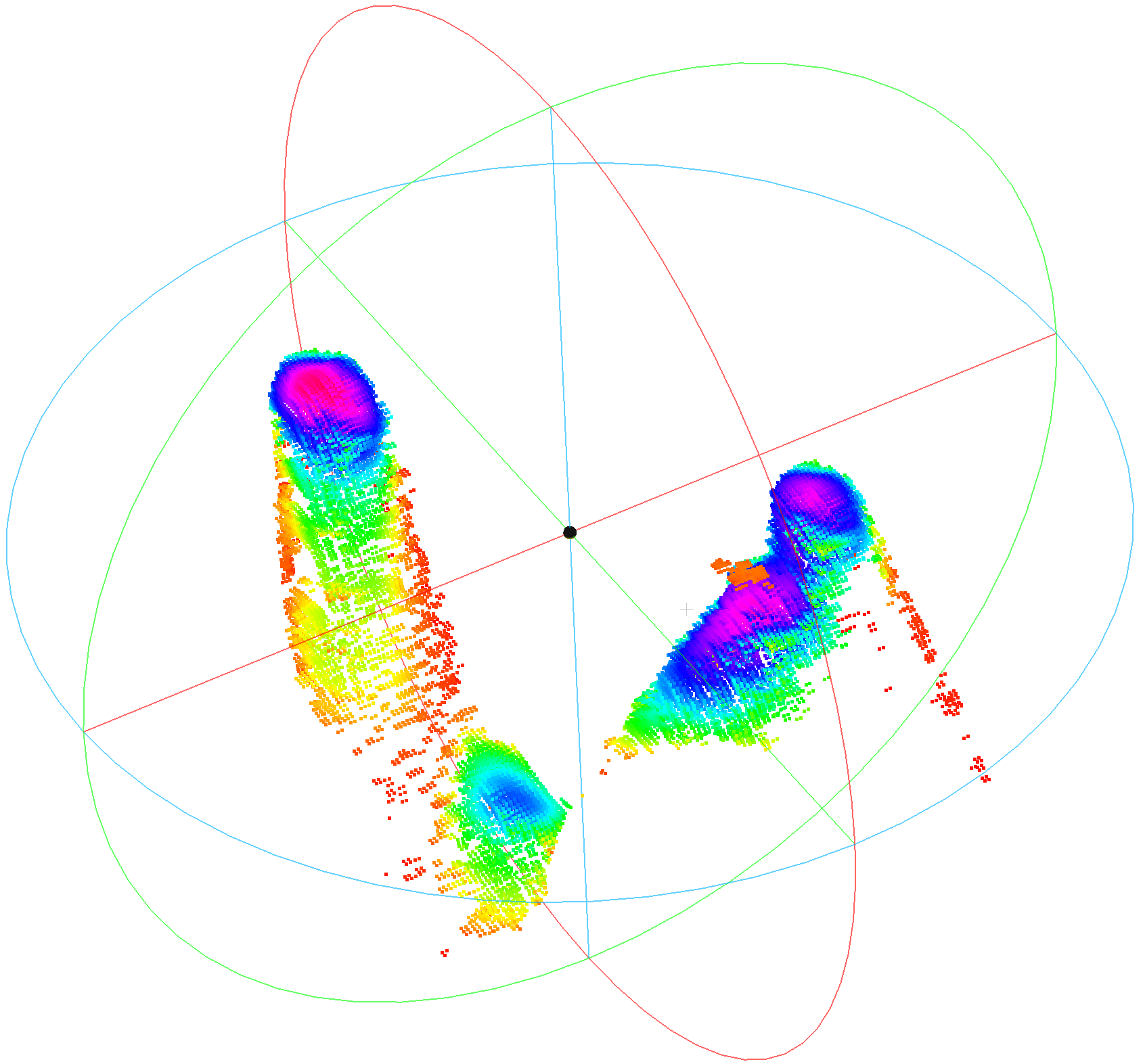}
     \end{subfigure}
     \begin{subfigure}[b]{0.155\textwidth}
         \centering
         \includegraphics[width=\textwidth]{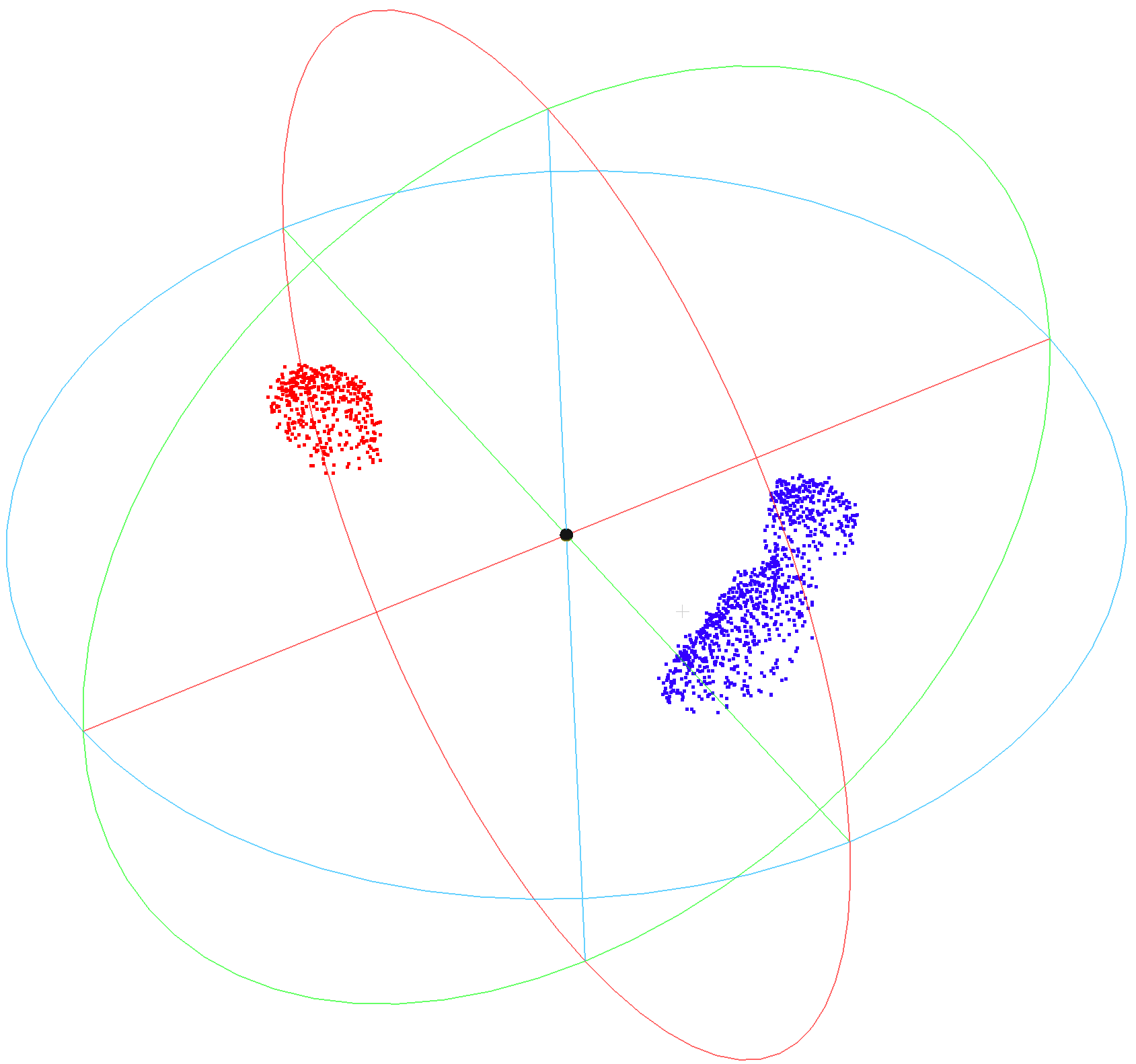}
     \end{subfigure}
    \caption{Point clouds of the objects of interest isolated from the background (left), density maps produced by KDE (center), and clusters identified by HDBSCAN (right).}
    \label{fig:detection_result}
\end{figure}

\paragraph{Density-based Filtering.}
The range between the sensor and the object directly affects the density of the captured point clouds, which in turn provides valuable information to the KDE algorithm. We observe that density peaks at the tip of each knife handle as illustrated in Figure~\ref{fig:detection_result}. Thus, we use percentile filtering to exclude outliers, primarily the body of the knife handles. This filtering step further enhances the separability of the point clouds, enhancing their suitability for unsupervised cluster analysis.

% MAE, RMSE result
\begin{table}
    \centering
    \begin{tabular}{crrr}
        \hline
        \textsc{Object(s)}  & MAE $\downarrow$ & RMSE $\downarrow$ & NRMSE $\downarrow$ \\
        \hline
        \hline
        1     &0.168     &0.410  &  0.410  \\
        2     &0.477     &0.924  &  0.462  \\
        3     &0.655     &1.263  &  0.421  \\
        4     &0.953     &1.520  &  0.380  \\
        5     &1.114     &1.876  &  0.375  \\
        \hline
    \end{tabular}
    \caption{Detection performance for scenarios with 1 to 5 objects. Metrics MAE, RMSE, and NRMSE indicate the average detection error, with lower values reflecting fewer misses per attempt.}
    \label{tab:result}
\end{table}

% Side-by-side heatmap and bbox
\begin{figure}[!b]
    \centering
    \begin{subfigure}[b]{0.15\textwidth}
        \centering
        \includegraphics[width=\textwidth]{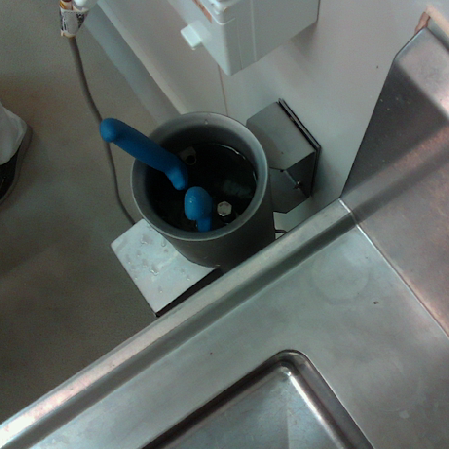}
    \end{subfigure}
    \begin{subfigure}[b]{0.15\textwidth}
        \centering
        \includegraphics[width=\textwidth]{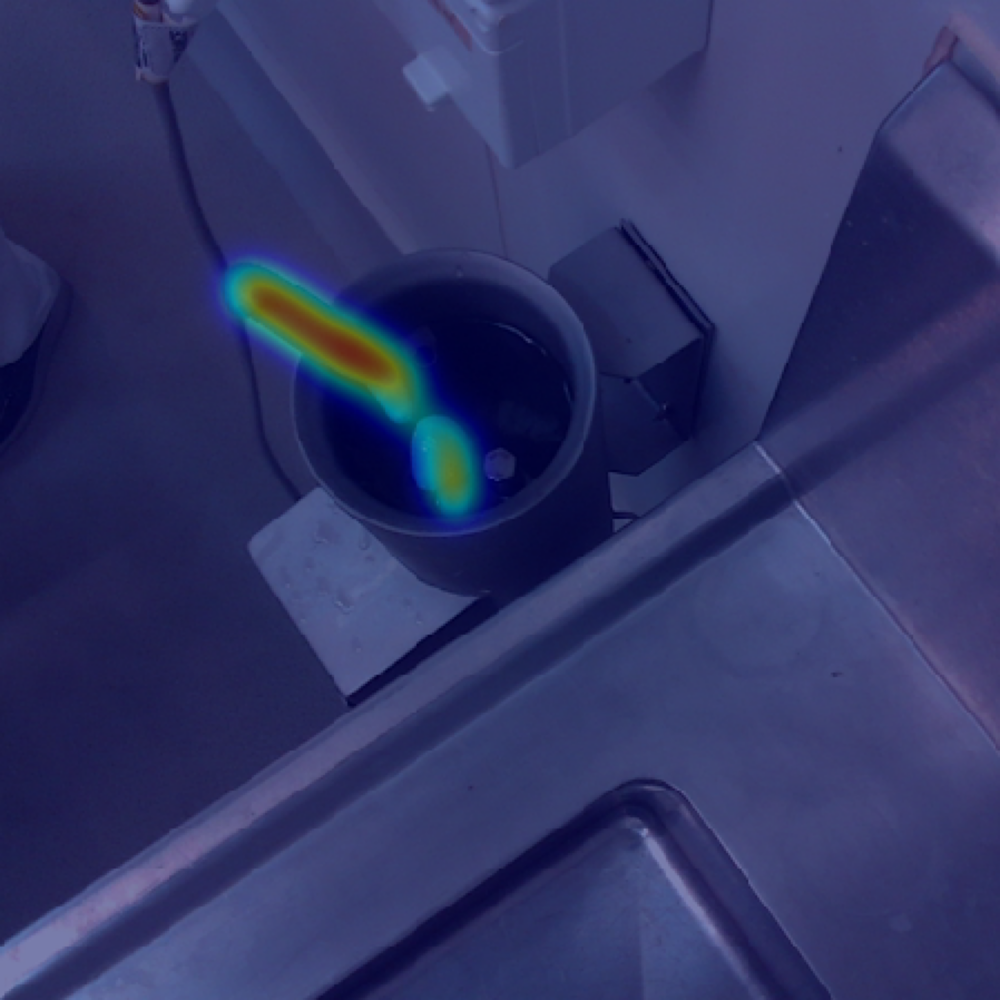}
    \end{subfigure}
    \begin{subfigure}[b]{0.15\textwidth}
        \centering
        \includegraphics[width=\textwidth]{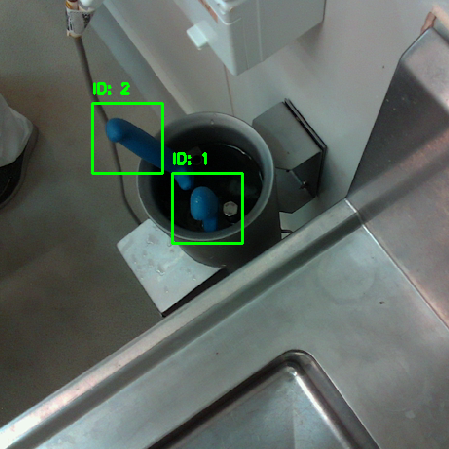}
    \end{subfigure}
    \caption{Illustration showing the original image (left), the 2D projected heatmap (center), and the detection results (right)}
    \label{fig:2d_heatmap_bbox}
\end{figure}

\paragraph{Detection Performance.}
The final detection results, produced by the SORT algorithm, take the form of bounding boxes. Figure~\ref{fig:2d_heatmap_bbox} presents a side-by-side view of a heatmap of the objects of interest alongside the corresponding detection results. To accurately quantify discrepancies between the system's predictions and the ground truth, we employ the mean absolute error (MAE), root mean square error (RMSE), and normalized RMSE (NRMSE) performance measures. Corresponding results for varying numbers of objects are comprehensively presented in Table~\ref{tab:result}.
The evaluation results demonstrate the commendable performance of our system, even though there is a slight increase in the detection error as the number of objects increases, which can primarily be attributed to the inherent challenges associated with point cloud clustering. 

\begin{figure}[t]
  \centering
  \includegraphics[width=0.46\textwidth]{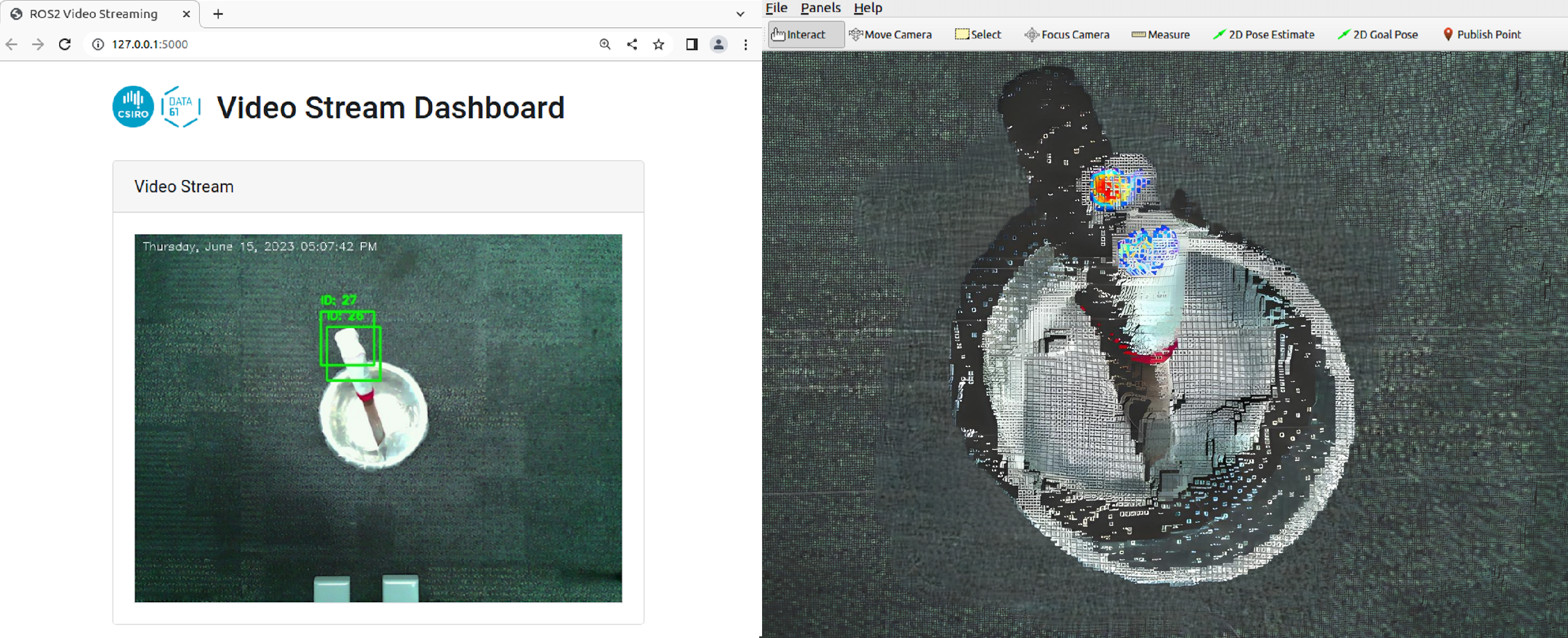}
  \caption{Left: A web-based visualization system implemented using a Flask API. Right: A real-time 3D visualization of point clouds overlaid with density maps using RViz from ROS 2.}
  \label{fig:web_ui}
\end{figure}

Compared to conventional color-based object detection approaches, our density-based detection pipeline is capable of separating closely clumped objects with similar colors. As illustrated in Figure~\ref{fig:web_ui}, our proposed method successfully distinguishes between two knife handles that are both white and positioned closely together, demonstrating the method's reliability and effectiveness.

To ensure regulatory compliance, the system not only detects the presence of a knife in the sanitation bath but also assigns a unique ID to each detected knife and tracks its duration within the bath. This tracking enables the system to determine whether each knife has been sanitized for a sufficient length of time. Once a knife meets the required sanitation duration, the system visually indicates compliance by changing the color of the bounding box from red to green, signaling that the knife is ready to be used.

\section{Conclusion}
We have developed a system using a Time-of-Flight (ToF) camera to ensure knife sanitation compliance by detecting and tracking knives in a sanitation bath. The system's potential extends beyond knife sanitation, being applicable in various industry quality assurance tasks. Additionally, its label-less approach holds promise for generating pseudo-labels for few-shot learning in object detection, expanding its utility in machine learning applications where labeled data is limited.

The system operates in real-time but faces limitations due to pre-processing steps like density estimation, which are inherently slow and can be disrupted by extraneous objects. Future work will explore more efficient 3D object detection architectures to overcome these challenges. Overall, the system offers a practical solution for ensuring compliance, with promising avenues for expanding its utility and enhancing its performance across various domains.

\newpage

\section*{Acknowledgements}
This work was partially supported by CSIRO's Science Leader project R-91559.
\bibliographystyle{named}
\bibliography{ijcai24}

\begin{thebibliography}{}

\bibitem[\protect\citeauthoryear{Bewley \bgroup \em et al.\egroup }{2016}]{DBLP:conf/icip/BewleyGORU16}
Alex Bewley, ZongYuan Ge, Lionel Ott, Fabio~Tozeto Ramos, and Ben Upcroft.
\newblock Simple online and realtime tracking.
\newblock In {\em 2016 {IEEE} International Conference on Image Processing, {ICIP} 2016, Phoenix, AZ, USA, September 25-28, 2016}, pages 3464--3468. {IEEE}, 2016.

\bibitem[\protect\citeauthoryear{Khamis \bgroup \em et al.\egroup }{2020a}]{DBLP:conf/sensys/KhamisKCMH20}
Abdelwahed Khamis, Branislav Kusy, Chun~Tung Chou, Mary{-}Louise McLaws, and Wen Hu.
\newblock Rfwash: a weakly supervised tracking of hand hygiene technique.
\newblock In Jin Nakazawa and Polly Huang, editors, {\em SenSys '20: The 18th {ACM} Conference on Embedded Networked Sensor Systems, Virtual Event, Japan, November 16-19, 2020}, pages 572--584. {ACM}, 2020.

\bibitem[\protect\citeauthoryear{Khamis \bgroup \em et al.\egroup }{2020b}]{DBLP:conf/ipsn/KhamisKCMH20}
Abdelwahed Khamis, Branislav Kusy, Chun~Tung Chou, Marylouise McLaws, and Wen Hu.
\newblock Poster abstract: {A} weakly supervised tracking of hand hygiene technique.
\newblock In {\em 19th {ACM/IEEE} International Conference on Information Processing in Sensor Networks, {IPSN} 2020, Sydney, Australia, April 21-24, 2020}, pages 331--332. {IEEE}, 2020.

\bibitem[\protect\citeauthoryear{McInnes \bgroup \em et al.\egroup }{2017}]{DBLP:journals/jossw/McInnesHA17}
Leland McInnes, John Healy, and Steve Astels.
\newblock hdbscan: Hierarchical density based clustering.
\newblock {\em J. Open Source Softw.}, 2(11):205, 2017.

\bibitem[\protect\citeauthoryear{St{\"{a}}cker \bgroup \em et al.\egroup }{2021}]{DBLP:conf/iccvw/StackerFHBRSS21}
Lukas St{\"{a}}cker, Juncong Fei, Philipp Heidenreich, Frank Bonarens, Jason~R. Rambach, Didier Stricker, and Christoph Stiller.
\newblock Deployment of deep neural networks for object detection on edge {AI} devices with runtime optimization.
\newblock In {\em {IEEE/CVF} International Conference on Computer Vision Workshops, {ICCVW} 2021, Montreal, BC, Canada, October 11-17, 2021}, pages 1015--1022. {IEEE}, 2021.

\bibitem[\protect\citeauthoryear{Wang \bgroup \em et al.\egroup }{2022}]{DBLP:conf/mobicom/WangLABMB22}
Ziwei Wang, Jiajun Liu, Reza Arablouei, Greg Bishop{-}Hurley, Melissa Matthews, and Paulo Borges.
\newblock Multi-modal sensing for behaviour recognition.
\newblock In {\em {ACM} MobiCom '22: The 28th Annual International Conference on Mobile Computing and Networking, Sydney, NSW, Australia, October 17 - 21, 2022}, pages 900--902. {ACM}, 2022.

\bibitem[\protect\citeauthoryear{Zhang \bgroup \em et al.\egroup }{2020}]{DBLP:journals/corr/abs-2006-10214}
Fan Zhang, Valentin Bazarevsky, Andrey Vakunov, Andrei Tkachenka, George Sung, Chuo{-}Ling Chang, and Matthias Grundmann.
\newblock Mediapipe hands: On-device real-time hand tracking.
\newblock {\em CoRR}, abs/2006.10214, 2020.

\end{thebibliography}

\end{document}